\documentclass[letterpaper, 10 pt, conference]{ieeeconf}

\IEEEoverridecommandlockouts 

\usepackage{cite}
\usepackage{amsmath, amssymb, amsfonts}
\usepackage{algorithmic}
\usepackage{graphicx}
\usepackage{subcaption} 
\usepackage{textcomp}
\usepackage{xcolor}
\usepackage{multirow}
\usepackage{url}
\usepackage[hidelinks]{hyperref}
\def\BibTeX{{\rm B\kern-.05em{\sc i\kern-.025em b}\kern-.08em T\kern-.1667em\lower.7ex\hbox{E}\kern-.125emX}}
\begin{document}
    \title{
    VoxelFormer: Parameter-Efficient Multi-Subject Visual Decoding from fMRI}

    \author{Chenqian Le$^{1}$, Yilin Zhao$^{1}$, Nikasadat Emami$^{1}$, 
    Kushagra Yadav$^{2}$, Xujin "Chris" Liu$^{1}$, Xupeng Chen$^{1}$, Yao Wang$^{1}$%
    \thanks{$^{1}$Department of Electrical and Computer Engineering, New York University Tandon School of Engineering, Brooklyn, NY, USA. {\tt\small \{cl6707, yz10381, ne2213, xl3942, xc1490, yaowang\}@nyu.edu}}%
    \thanks{$^{2}$Department of Computer Science and Engineering, New York University Tandon School of Engineering, Brooklyn, NY, USA. {\tt\small ky2684@nyu.edu}}}

    \maketitle

    \begin{abstract}
        Recent advances in fMRI-based visual decoding have enabled compelling reconstructions
        of perceived images. However, most approaches rely on subject-specific training,
        limiting scalability and practical deployment. We introduce \textbf{VoxelFormer},
        a lightweight transformer architecture that enables multi-subject training
        for visual decoding from fMRI. VoxelFormer integrates a Token Merging
        Transformer (ToMer) for efficient voxel compression and a query-driven Q-Former
        that produces fixed-size neural representations aligned with the CLIP
        image embedding space. Evaluated on the 7T Natural Scenes Dataset,
        VoxelFormer achieves competitive retrieval performance on subjects
        included during training with significantly fewer parameters than
        existing methods. These results highlight token merging and query-based transformers
        as promising strategies for parameter-efficient neural decoding. The
        source code is available at
        \url{https://github.com/kushagrayadv/voxel-former}.
    \end{abstract}


    \begin{figure*}[t]
        \includegraphics[width=\linewidth]{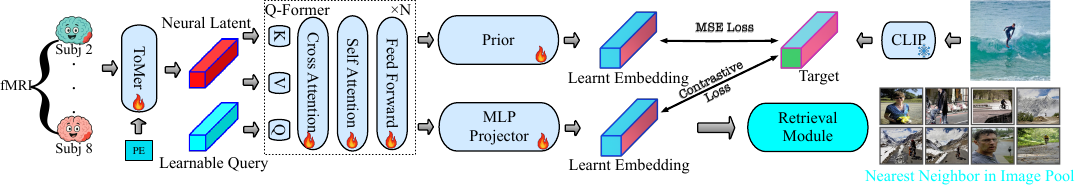}
        \caption{Overview of the proposed \textbf{VoxelFormer} pipeline for cross-subject
        fMRI-to-image decoding. Multi-subject fMRI volumes are first encoded
        using a token-merging encoder (\textbf{ToMer}) with coordinate-based positional
        embeddings (PE) to generate compact neural latents. These latents
        interact with a small set of learnable queries within a Query-Former (\textbf{Q-Former}),
        composed of repeated cross-attention, self-attention, and feed-forward
        layers, to produce subject-invariant embeddings. The resulting
        representation branches into two decoding heads: (i) a Prior transformer
        trained with mean squared error (MSE) loss to regress frozen CLIP image embeddings,
        and (ii) an MLP projector trained with a contrastive loss for image
        retrieval via a nearest-neighbor search module. Trainable modules are denoted
        with a fire symbol, while the CLIP encoder remains frozen. In this work,
        only the retrieval branch is evaluated.}
        \label{fig:pipeline}
    \end{figure*}
    \section{Introduction}

    Decoding human visual perception from fMRI signals can transform brain–computer
    interfaces, clinical neuroimaging, and our understanding of how the visual cortex
    encodes complex scenes\cite{yamins_using_2016, glaser_machine_2020,naselaris_encoding_2011
    }. Most accurate decoders to date rely on massive subject-specific datasets or
    require extensive anatomical or functional alignment, hindering scalability
    and limiting practical deployment.

    In this work, we ask: \textbf{Can we build a parameter-efficient visual
    decoder that leverages multi-subject training data effectively?} To address this,
    we propose \textbf{VoxelFormer}, a two-stage transformer architecture that (1)
    compresses and fuses raw voxel activations into a compact latent
    representation using a novel \textbf{ToMer encoder}, and (2) refines these
    features via a \textbf{Q-Former} to align with the CLIP image embedding\cite{radford_learning_2021}
    space while producing fixed-size representations across subjects.

    We evaluate on the 7T Natural Scenes Dataset (NSD)\cite{allen2021naturalscenes}
    across eight participants. VoxelFormer achieves competitive performance for the subjects seen during training as well as lower parameter counts compare to other works.

    Our main contributions are:
    \begin{enumerate}
        \item A \textbf{Token Merging Transformer (ToMer)} that dynamically
            reduces the fMRI token count via learned attention, lowering memory cost
            while preserving critical information.

        \item A \textbf{query-driven Q-Former} that produces fixed-size latent
            representations enabling multi-subject training.

        \item Demonstration of parameter-efficient multi-subject visual decoding
            achieving competitive performance with significantly reduced model size.
    \end{enumerate}
    \section{Related Work}

    \subsection{Subject-Specific Visual Decoding}
    Early fMRI-based visual decoders map each subject’s voxel activations to image
    or feature representations. Shen \emph{et al.}~\cite{shen2019deep} trained deep
    generative models per subject to reconstruct images, requiring hundreds of
    images per individual. Scotti \emph{et al.}~\cite{scotti2023mindeye}
    demonstrated image retrieval by fine-tuning a subject-specific  encoder on CLIP features \cite{CLIP}, achieving strong within-subject performance at the cost of per-user adaptation. The resulting model is referred to as MindEye1. MindEye1 maps ~15000 visual-cortex voxels directly to the full 257 $\times$ 768 CLIP token matrix using a 4-block residual MLP that contains $\approx$ 940M parameters-over an order of magnitude larger than earlier linear or shallow-network decoders. Because this massive network is trained independently for each participant, it must see tens of thousands of stimulus-voxel pairs (30-40 hours of scans in the NSD) to avoid over-fitting to idiosyncratic voxel partterns and to learn a stable voxel-to-token mapping.

    \subsection{Cross-Subject Alignment}
    To mitigate the need for per-subject data, alignment methods project
    multiple brains into a shared space. Hyperalignment~\cite{shen2019deep} align voxel responses across individuals with co-registration. While anatomical
    alignment is routinely performed in clinical MRI and may be necessary for
    meaningful multi-subject training (since the same voxel coordinate does not represent
    the same brain location across individuals), more recent networks~\cite{scotti2024mindeye2,wang_mindbridge_2024} focus on improving generalization through architectural innovations. The MindEye2 framework~\cite{scotti2024mindeye2} uses subject-specific layers (a ridge regession layer) to map raw fMRI data from different training subjects into a common latent space, which is then transformed using a four-block residual MLP backbone to the features similar to CLIP features for the same input image. After pre-training the shared pipeline on seven NSD subjects ( $\approx$ 250h of data in total), MindEye2 fine-tunes the entire model with as little as 1 hour of fMRI from a new subject, yet attains reconstruction quality comparable to a single-subject MindEye1 model trained on the full 40 hours scan set.
    

    \subsection{Transformer-Based Token Compression \& Query Encoders}
    Token-merging techniques for vision transformers, such as Token Merging (ToMe)~\cite{bolya2022tome}
    and Tokens-to-Token (T2T)~\cite{yuan2021tokens}, reduce compute by merging redundant
    patches via attention. Perceiver~\cite{jaegle_perceiver_2021} and Q-Former~\cite{li2023blip2}
    architectures use learned queries to distill variable-size inputs into a
    fixed latent. However, these strategies have not been fully explored for
    fMRI decoding across subjects.

    \emph{VoxelFormer} integrates dynamic token merging and query-based encoding
    to achieve parameter-efficient multi-subject training for neural decoding.
    \section{Method}

    \subsection{Dataset}
    We use the 7T Natural Scenes Dataset (NSD)~\cite{allen2021naturalscenes},
    which comprises whole-brain, high-resolution fMRI from eight adults, each
    exposed to thousands of natural scene images from Microsoft COCO~\cite{lin_microsoft_2015}
    over 30–40 sessions. This dataset
    is well-suited for evaluating non-invasive brain-based visual decoding. In our case, we use S2-S7 together to train the model.

    \subsection{ToMer Encoder}
    \begin{figure}[h]
        \centering
        \includegraphics[width=\linewidth]{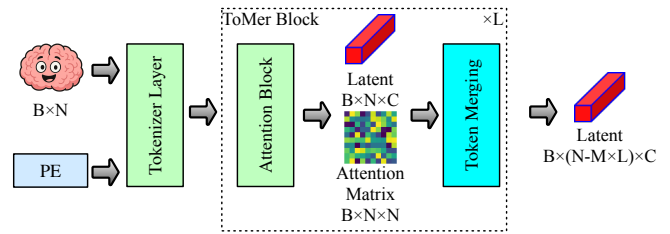}
        \caption{\textbf{ToMer Encoder Architecture} The ToMer encoder processes
        input fMRI data ($B×N$, where $B$ is batch size and $N$ is the number of voxels
        in the visual cortex) by first applying a Tokenizer Layer and Positional
        Embedding (PE). The tokenized features are then passed through an attention
        block, producing latent representations ($B×N×C$) and an attention matrix
        ($B×N×N$). A Token Merging operation reduces the number of tokens by merging
        those with the highest attention similarity. This encoder structure can
        be stacked $L$ times and serves as the neural feature extractor in our
        VoxelFormer framework. If $M$ tokens are reduced in each stage, the
        compressed latent representation has a shape B×(N–M×L)×C}
        \label{fig:tomer}
    \end{figure}

    We introduce a novel Transformer-based encoder called \textbf{Token Merging
    Transformer (ToMer)} to efficiently process high-dimensional fMRI data. As
    depicted in Fig.~\ref{fig:tomer}, ToMer first tokenizes the input voxel data
    using a 1$\times$1 convolutional layer, followed by the addition of sinusoidal positional
    embeddings derived from voxel coordinates via a SiREN\cite{sitzmann2019siren}
    module. Subsequently, a self-attention mechanism captures the relationships
    among tokens, yielding latent neural representations and corresponding attention
    matrices.

    Leveraging the learned attention scores, the ToMer encoder dynamically merges pairs of highly correlated tokens through the Token Merging operation~\cite{bolya2022tome}. In the original ToMe formulation, this attention-guided merging is applied only at inference time to accelerate forward passes by reducing the effective token count. In contrast, we integrate Token Merging directly into the training loop, merging tokens on-the-fly as gradients propagate. This yields a compressed, progressively coarsened representation throughout both learning and evaluation, substantially lowering computational and memory complexity during training without sacrificing task performance.

    The ToMer block can be stacked multiple times, progressively condensing the neural
    representation into a compact and informative latent space. This adaptive
    compression strategy is critical for enabling scalable and efficient decoding across subjects, making it well-suited for the cross-subject fMRI decoding
    problem tackled by our VoxelFormer model.

    \subsection{Q-Former}
    To enable robust multi-subject training and produce fixed-size representations
    regardless of the original number of voxels, we propose to use \cite{Q-former},
    a query-based transformer module that produces consistent-sized neural
    embeddings (Fig.~\ref{fig:pipeline}). Specifically, the Q-Former accepts compressed
    neural latent features from the ToMer encoder and utilizes learnable queries
    to flexibly attend and aggregate the most salient information from individual
    subject's brain data into a representation similar to a chosen learnt feature space. In our case, we used the CLIP features \cite{CLIP}.


    Specifically, the Q-Former employs a cross-attention mechanism where a fixed set of trainable
    query tokens repeatedly attend to the variable number of token features produced by the ToMer. The resulting embeddings from the query tokens at the last stage provide a consistent representation
    size that facilitates multi-subject training and alignment with visual features
    in the CLIP embedding space.

Note that the proposed pipeline in Fig.~\ref{fig:pipeline}) enables training with multiple subject data without using subject-specific layers as in MindEye2.

      \subsection{Loss Function}
    To facilitate stable training and ensure compatibility with downstream tasks,
    we follow a dual-pathway training strategy inspired by recent work~\cite{scotti2024mindeye2}.
    The output embeddings from the Q-Former branch into two distinct modules: (1)
    a \emph{prior transformer}, which aligns embeddings to CLIP-derived visual
    embeddings via mean squared error (MSE) loss, theoretically enabling potential
    use as conditioning signals for diffusion-based image generation ; and (2) an
    \emph{MLP projector}, trained with a contrastive loss, which directly
    supports robust image retrieval from a visual database through nearest-neighbor
    search. While our current experiments focus primarily on image retrieval performance,
    the prior transformer pathway could potentially be used for image reconstruction using a diffusion model. Training both branches together enables the shared modules (the ToMer and Q-Former) to produce good features for both branches.

    Specifically, we use a two‐phase training schedule: for the first one‐third of the epochs, the MLP projector is trained with the BiMixCo contrastive loss (combined with MSE on the prior branch), and for the remaining two‐thirds of training we replace BiMixCo with the SoftCLIP loss (while continuing to optimize the MSE term).

    \textbf{Mean Squared Error (MSE) Loss}: This loss aligns the prior transformer
    embeddings $\mathbf{z}_{i}^{\text{prior}}$ with the CLIP-generated visual embeddings
    $\mathbf{z}_{i}^{\text{CLIP}}$ for the same $i$-th image stimulus using the MSE loss formula:
    \begin{align}
        \mathcal{L}_{\text{MSE}}= \frac{1}{B}\sum_{i=1}^{B}\|\mathbf{z}_{i}^{\text{prior}}- \mathbf{z}_{i}^{\text{CLIP}}\|^{2}
    \end{align}

    \textbf{Contrastive Loss with MixCo and Soft CLIP}: The MLP projector embeddings
    $\mathbf{z}_{i}^{\text{MLP}}$ utilize a combination of InfoNCE contrastive loss
    and Mixup data augmentation, collectively referred to as BiMixCo. We further combine
     BiMixCo loss with  SoftCLIP loss to enhance the
    discriminative power of the embeddings, while aligning the resulting features with the CLIP features. 

    \textbf{BiMixCo Loss} is defined as:
    \begin{align}
        \mathcal{L}_{\text{BiMixCo}}= - \sum_{i=1}^{N}\left[ \lambda_{i}\cdot \log \left( \frac{\exp\left( \frac{p_{i}^{*}\cdot t_{i}}{\tau} \right)}{\sum_{m=1}^{N}\exp\left( \frac{p_i^* \cdot t_m}{\tau}\right)}\right) \right. \nonumber \\
        \left. + (1 - \lambda_{i}) \cdot \log \left( \frac{\exp\left( \frac{p_{i}^{*}\cdot t_{k_i}}{\tau} \right)}{\sum_{m=1}^{N}\exp\left( \frac{p_i^* \cdot t_m}{\tau}\right)}\right) \right] \nonumber                                    \\
        - \sum_{j=1}^{N}\left[ \lambda_{j}\cdot \log \left( \frac{\exp\left( \frac{p_{j}^{*}\cdot t_{j}}{\tau} \right)}{\sum_{m=1}^{N}\exp\left( \frac{p_m^* \cdot t_j}{\tau}\right)}\right) \right. \nonumber                               \\
        \left. + \sum_{\{l|k_l = j\}}(1 - \lambda_{l}) \cdot \log \left( \frac{\exp\left( \frac{p_{l}^{*}\cdot t_{j}}{\tau} \right)}{\sum_{m=1}^{N}\exp\left( \frac{p_m^* \cdot t_j}{\tau}\right)}\right) \right]
    \end{align}

    \textbf{SoftCLIP Loss} is defined as:
    \begin{align}
        \mathcal{L}_{\text{SoftCLIP}}= - \sum_{i=1}^{N}\sum_{j=1}^{N}\left[ \frac{\exp\left( \frac{t_{i}\cdot t_{j}}{\tau} \right)}{\sum_{m=1}^{N}\exp\left( \frac{t_i \cdot t_m}{\tau}\right)}\right. \nonumber \\
        \left. \cdot \log \left( \frac{\exp\left( \frac{p_{i}\cdot t_{j}}{\tau} \right)}{\sum_{m=1}^{N}\exp\left( \frac{p_i \cdot t_m}{\tau}\right)}\right) \right]
    \end{align}
    Here, $p_i = \lambda_i\,\mathbf z_{i}^{\mathrm{MLP}} + (1-\lambda_i)\,\mathbf z_{k_i}^{\mathrm{MLP}}$ denotes the (possibly mixup-augmented) MLP-projector output for sample $i$; 
$t_i = \mathbf z_{i}^{\mathrm{CLIP}}$ is the frozen CLIP image embedding for the same sample; 
and $t_m = \mathbf z_{m}^{\mathrm{CLIP}}$ ($m=1,\ldots,N$) are all CLIP embeddings in the batch used as negatives (including $m=i$ in the denominator of each softmax).

    \textbf{Total Loss}: The overall training objective is expressed as:
    \begin{align}
        \mathcal{L}_{\text{total}}= \lambda_{\text{MSE}}\mathcal{L}_{\text{MSE}}+ \lambda_{\text{contrastive}}\mathcal{L}_{\text{contrastive}}
    \end{align}

    Here, $\lambda_{\text{MSE}}$ and $\lambda_{\text{contrastive}}$ are
    hyperparameters that balance the trade-off between reconstruction alignment
    and discriminative retrieval performance. In our case, we use
    $\lambda_{\text{MSE}}=30$ and $\lambda_{\text{contrastive}}=1$. This dual-phase
    loss strategy, beginning with MixCo and transitioning to Soft CLIP loss after training for the first $\frac{1}{3}$ epochs, supports
    robust generalization and stable model convergence, facilitating both
    effective zero-shot retrieval and potential image reconstruction.

    \section{Results}


    We evaluate VoxelFormer on the 7T NSD, following the standard top-1 retrieval
    protocol~\cite{scotti2023mindeye} with a candidate pool of 300 images.
    Image retrieval is performed by computing cosine similarity between brain-derived
    embeddings and CLIP image embeddings, then selecting the closest match.
    Forward retrieval measures accuracy when using brain embeddings to retrieve
    the correct image from the pool, while backward retrieval measures accuracy when
    using image embeddings to retrieve the correct brain response. Chance level
    performance is 0.33\% (1/300).

\paragraph{Subject-Wise Performance}
    Table~\ref{tab:retrieval_subjectwise} summarizes retrieval accuracy for
    individual subjects. For our model, 7 subjects were used during training (subjects 2-7),
    with subject 1 was held out for evaluation of zero-shot retrieval performance.
    Compared to MindEye1 and MindEye2, which are state-of-the-art subject-specific
    and aligned models with substantially larger parameter counts, VoxelFormer
    achieves competitive performance on subjects included in the training set.   It is important to recall that MindEye1 has a separate trained model for each subject, whereas MindEye2 trained a single model using subjects 2-7, with subject specific input layers plus a shared module. VoxFormer has a shared ToMer module and a shared Q-Former module, without any subject-specific layers.
    Despite the absence of the subject-specific layers,  VoxelFormer achieved consistently
    above 66\% accuracy on all evaluated subjects, underlining the robustness of our query-based
    representation.
    \begin{table}[h]
        \centering
        \renewcommand{\arraystretch}{1.2}
        \caption{Top-1 retrieval accuracy (\%) by subject for Subjects within
        Training Data.}
        \begin{tabular}{lccc}
            \hline
            \textbf{Subject}   & \textbf{Method} & \textbf{Fwd Acc. (\%)} & \textbf{Bwd Acc. (\%)} \\
            \hline
            \multirow{3}{*}{2} & MindEye1        & 97.1                   & 93.9                   \\
                               & MindEye2        & 99.88                  & 99.84                  \\
                               & Ours            & 86.54                  & 85.78                  \\
            \hline
            \multirow{2}{*}{3} & MindEye1        & 90.7                   & 85.7                   \\
                               & Ours            & 74.97                  & 74.17                  \\
            \hline
            \multirow{2}{*}{4} & MindEye1        & 89.4                   & 85.9                   \\
                               & Ours            & 75.15                  & 73.36                  \\
            \hline
            \multirow{2}{*}{5} & MindEye2        & 98.39                  & 96.94                  \\
                               & Ours            & 73.03                  & 71.62                  \\
            \hline
            \multirow{1}{*}{6} & Ours            & 74.93                  & 74.16                  \\
            \hline
            \multirow{2}{*}{7} & MindEye2        & 96.89                  & 96.53                  \\
                               & Ours            & 68.65                  & 67.46                  \\
            \hline
        \end{tabular}
        \label{tab:retrieval_subjectwise}
    \end{table}


    \begin{table}[h]
        \centering
        \renewcommand{\arraystretch}{1.2}
        \caption{Mean retrieval accuracy (Top-1) across training subjects and Model Size. 
        }

        \begin{tabular}{lccc}
            \hline
            \textbf{Method}                                                & \textbf{Fwd Acc. (\%)} & \textbf{Bwd Acc. (\%)} & \textbf{Param} \\
            \hline
            MindEye1\cite{scotti2023mindeye} (S1-S4)                               & 93.6                   & 90.1                   & 940M           \\
            MindEye2\cite{scotti2024mindeye2} (S1,2,5,7)                             & \textbf{98.3}          & \textbf{98.3}          & 469M           \\
            Brain Diffuser\cite{ozcelik2023naturalscenereconstructionfmri} (S1-S4) & 21.1                   & 30.3                   & --             \\
            Ours (S2–S7)                                                   & 74.3                   & 73.1                   & \textbf{39M}   \\
            \hline
        \end{tabular}

        \label{tab:retrieval_mean}
    \end{table}



    \paragraph{Parameter Efficiency}
    A major contribution of VoxelFormer is its parameter efficiency.  Table~\ref{tab:retrieval_mean} compares mean top-1 retrieval accuracy across
    training subjects for all methods, along with the model size. 
    While MindEye2 achieves the highest mean retrieval accuracy, it uses over 469M parameters (counting one subject-specific layer for the S1, plus the shared module). MindEye2 achieves slightly worse performance, even though each subject specific model is twice the size the MindEye2.  
    In contrast, VoxelFormer attains reasonable mean accuracy  with only 39M parameters—a 12$\times$ reduction in model size
    compared to MindEye2 and 24$\times$ reduction when compared to MindEye1. This
    demonstrates that our proposed architecture can achieve competitive performance
    with significantly fewer parameters.
    Our findings suggest that architectural design—specifically
    attention-guided token merging and query-based feature distillation—can
    compensate for reduced capacity, offering an efficient path forward for future
    neural decoders, particularly in resource-constrained settings.

    \section{Discussion}

    We present \textbf{VoxelFormer}, a lightweight transformer framework that
    combines token-merging for voxel compression with a query-driven alignment
    module, enabling parameter-efficient multi-subject visual decoding from fMRI.
    While retrieval accuracy remains below state-of-the-art subject-specific
    approaches, VoxelFormer demonstrates that competitive performance can be achieved
    with significantly fewer parameters through careful architectural design.

    Crucially, VoxelFormer is far more compact than recent baselines—39M
    parameters versus over 469M in MindEye2—while remaining competitive for
    subjects included in training. This demonstrates that token reduction and query-based
    transformers are promising strategies for parameter-efficient neural decoders that can be trained using data for multiple subjects.

    Future work will explore improved cross-subject architectures, anatomical
    alignment strategies, larger pretraining datasets, and joint optimization
    for image reconstruction, with the goal of further closing the performance gap
    while maintaining parameter efficiency. VoxelFormer provides a foundation
    for parameter-efficient neural decoding that could be valuable in resource-constrained
    settings.

    \bibliographystyle{IEEEtran}
    \bibliography{reference}
    \newpage


\end{document}